\documentclass[sigconf]{acmart}

\AtBeginDocument{%
  }

\setcopyright{none}

\usepackage{todonotes}
\usepackage{subcaption}
\usepackage{graphicx}
\usepackage{multirow}
\usepackage{multicol}
\usepackage{tabularx}
\usepackage{makecell}
\usepackage{xcolor}

\begin{document}

\title{Evaluating graph-based explanations for AI-based recommender systems}

\author{Simon Delarue}
\email{simon.delarue@telecom-paris.fr}
\affiliation{%
  \institution{LTCI, Télécom Paris, \\ Institut Polytechnique de Paris}
  \streetaddress{}
  \city{}
  \state{}
  \country{France}
  \postcode{}
}

\author{Astrid Bertrand}
\affiliation{%
  \institution{LTCI, Télécom Paris, \\ Institut Polytechnique de Paris}
  \streetaddress{}
  \city{}
  \state{}
  \country{France}
  \postcode{}
}

\author{Tiphaine Viard}
\affiliation{%
  \institution{i3, Télécom Paris, \\ Institut Polytechnique de Paris}
  \streetaddress{}
  \city{}
  \state{}
  \country{France}
  \postcode{}
}

\renewcommand{\shortauthors}{}

\newcommand{\ra}{\rightarrow}

\begin{abstract}
Recent years have witnessed a rapid growth of recommender systems, providing suggestions in numerous applications with potentially high social impact, such as health or justice. Meanwhile, in Europe, the upcoming AI Act mentions \emph{transparency} as a requirement for critical AI systems in order to ``mitigate the risks to fundamental rights''. Post-hoc explanations seamlessly align with this goal and extensive literature on the subject produced several forms of such objects, graphs being one of them. Early studies in visualization demonstrated the graphs' ability to improve user understanding, positioning them as potentially ideal explanations. However, it remains unclear how graph-based explanations compare to other explanation designs. In this work, we aim to determine the effectiveness of graph-based explanations in improving users' perception of AI-based recommendations using a mixed-methods approach. We first conduct a qualitative study to collect users' requirements for graph explanations. We then run a larger quantitative study in which we evaluate the influence of various explanation designs, including enhanced graph-based ones, on aspects such as understanding, usability and curiosity toward the AI system. We find that users perceive graph-based explanations as more usable than designs involving feature importance. However, we also reveal that textual explanations lead to higher objective understanding than graph-based designs. Most importantly, we highlight the strong contrast between participants' expressed preferences for graph design and their actual ratings using it, which are lower compared to textual design. These findings imply that meeting stakeholders' expressed preferences might not alone guarantee ``good'' explanations. Therefore, crafting hybrid designs successfully balancing social expectations with downstream performance emerges as a significant challenge.


\end{abstract}

\settopmatter{printacmref=false} 
\settopmatter{printccs=false}    
\renewcommand\footnotetextcopyrightpermission[1]{} 
\pagestyle{plain} 

\maketitle

\section{Introduction}

Recommender systems have emerged as fundamental tools for delivering personalized services to users. These frameworks find application across diverse domains, ranging from commonplace ones like online library and e-shopping, to more contentious ones such as finance, law, education or health. In cases where recommendations carry significant implications for users, it becomes essential to provide additional explanatory mechanisms. In Europe, ongoing legal discussions highlight the likelihood of transparency becoming a requirement for high-risk Artificial Intelligence (AI) systems~\cite{AIACT}. From the academic perspective, several works underscore a positive correlation between users' understanding of the model and their trust in such systems~\cite{sinha2002role,tintarev2007explanations,wilkenfeldFunctionalExplainingNew2014,zhangExplainableRecommendationSurvey2020,langerWhatWeWant2021}.

Explanations aim to address the \emph{``why''} question in relation to a recommender system or a specific prediction~\cite{zhangExplainableRecommendationSurvey2020}. Emerging from two primary research fields, namely Computer Science (CS) and Human Computer Interaction (HCI), various forms of explanations, such as text~\cite{wangGraphbasedExtractiveExplainer2022}, charts~\cite{herlockerExplainingCollaborativeFiltering2000,koukiUserPreferencesHybrid2017}, matrices~\cite{chen2018sequential}, hybrid designs~\cite{lundberg2017unified,digiacomoUserStudyHybrid2021,bertrand2023questioning} or graphs~\cite{wang2019explainable,popeExplainabilityMethodsGraph2019,yingGNNExplainerGeneratingExplanations}, have been proposed to enhance user satisfaction toward AI systems. Nevertheless, determining whether an explanation qualifies as a ``good'' one is not straightforward. On the one hand, Miller~\cite{miller2019explanation} argues that ``good'' explanations should be (i) \emph{contrastive}, highlighting contrast with alternatives, (ii) \emph{selected}, recognizing that people seldom expect the complete cause of an event and (iii) preferring \emph{causal} over probabilistic reasoning. Meanwhile, authors in~\cite{langerWhatWeWant2021} suggest that it is essential to ensure that explanations align with \emph{stakeholders' desiderata}, sometimes referred to as \emph{social expectations}~\cite{hilton1990conversational}.

\medskip

We hypothesize that graphs, as objects that can model selected relational data and exhibit causality, appears to align intuitively with Miller's criteria and may have the ability to fulfill various users' requirements. More specifically, the recommendation task can be naturally framed as a \emph{link prediction} task in a bipartite graph (see Figure~\ref{fig:graph_based_explanation_50951843}) that includes both users and items. The relevance of graph structures in problem-solving is not novel, as its roots trace back hundreds of years to Euler's solution to the Königsberg bridges problem~\cite{euler}. More recently, within the HCI field, several works have explored the impact of graph visualization through extended user studies~\cite{herman2000graph,burch2020state}. However, only a few works~\cite{koukiUserPreferencesHybrid2017,ghazimatinPRINCEProvidersideInterpretability2020} have explored the application of such designs in the explanatory context of recommender systems. In the CS community, when graphs are considered as explanations, they are often evaluated from an algorithmic perspective, also referred to as \emph{functionally-grounded evaluation}~\cite{doshi2018considerations}. This setup involves the use of machine learning-oriented proxy metrics that operate without human oversight, contradicting the criteria outlined in~\cite{hilton1990conversational,langerWhatWeWant2021}.

\medskip

In this work, we seek to bridge the gap between the HCI field, where graph designs are seldom examined in the context of explainability, and the CS field, where graphs are ubiquitous in explainable recommender systems~-- either as components of models or explanation designs~-- but are primarily evaluated from an algorithmic standpoint. To achieve this, we conduct a qualitative user study to characterize users' needs in terms of graph-based designs. Specifically, we gather requirements from users with varying levels of expertise in AI systems. Leveraging this knowledge, we develop an enhanced graph-based design and compare it to two commonly used forms of explanations~-- textual and SHAP~\cite{lundberg2017unified}-based explanations~-- through a quantitative user study. Through these studies, our goal is to answer the following research questions:




\begin{itemize}
    \item \textbf{RQ1}: \emph{What are the different stakeholders' desiderata regarding graph-based explanations for recommender systems?}
    \item \textbf{RQ2}: \emph{How does enhanced graph-based explanation design influence users of AI recommendations?}
    \item \textbf{RQ3}: \emph{How does enhanced graph-based explanation design compare to other explanation designs?}
    \item \textbf{RQ4}: \emph{How does the user's expertise level toward AI systems affect their explanation design preferences?}
\end{itemize}

\medskip

Our contributions encompass the following key aspects. Firstly, we investigate stakeholders' desiderata for graph-based explanations through a qualitative user study involving participants with diverse expertise levels, highlighting the preference for item-based design over user-based design. Secondly, drawing upon these insights, we build an enhanced graph-based explanation using item-oriented graph projection. Lastly, we conduct a larger scale quantitative study to discuss the impact of visual design on understanding, curiosity and usability. Consequently, we emphasize the contrast between users' expressed preference, both qualitatively and quantitatively, for graph-based explanations and their higher ratings when using dialogic explanations.



\section{Related work}

\subsection{Explainable AI for recommender systems}

Explainable recommendations offer user personalized item suggestions while explicitly clarifying \emph{why} a particular item is being proposed. The design of explanations for recommendations has been studied for a long time~\cite{herlockerExplainingCollaborativeFiltering2000,sinha2002role}. Numerous studies emphasized the correlation between understanding the system and placing trust in it~\cite{sinha2002role,tintarev2007explanations,wilkenfeldFunctionalExplainingNew2014,zhangExplainableRecommendationSurvey2020,langerWhatWeWant2021}, amplifying the significance of such explanations. While recommender systems can be crafted to naturally include interpretable elements (model-intrinsic)~\cite{zhang2014explicit}, our focus is on approaches involving \emph{post-hoc} explanations (model-agnostic), also known as post-hoc interpretability~\cite{liptonMythosModelInterpretability2018,miller2019explanation}.

Designing such explanation is a task situated at the intersection of two research fields, namely Computer Science (CS) and Human Computer Interaction (HCI), each with its distinct focus and requirements.

\subsection{Explainable AI in Computer Science}

From the CS perspective, building explainable recommendations involves a variety of techniques and explanation types. This includes rule mining~\cite{peake2018explanation}, approximation using simple models as with LIME~\cite{ribeiro2016should}, feature importance methods such as SHAP~\cite{lundberg2017unified}, attention maps~\cite{selvaraju2017grad} or graphs~\cite{wang2019explainable,popeExplainabilityMethodsGraph2019,yingGNNExplainerGeneratingExplanations}. Concerning graphs, authors often highlight their natural ability to reveal user-item connectivity~\cite{wang2019explainable} or extract small subgraphs containing elements most influential for predictions~\cite{yingGNNExplainerGeneratingExplanations}. In this context, justifications for using graphs as explanations for recommendations seem to align seamlessly with Miller's description of good explanations~\cite{miller2019explanation}.

However, the evaluation of such explanations in the CS field heavily relies on \emph{functionally-grounded} approaches~\cite{doshi2018considerations}, \emph{i.e.} algorithmic-oriented techniques that measure proxy metrics such as \emph{fidelity} or \emph{correctedness}~\cite{nauta2023anecdotal}, but lack human control. While such an evaluation approach offers advantages by saving time and avoiding ethical concerns potentially associated with human participation, authors in~\cite{miller2017explainable,doshi2018considerations} emphasize the limitations of these approaches in terms of ``real-world impact'' and advocate for their use solely after conducting user studies. To tackle this issue, recent work integrated both graph-based explanations and an evaluation framework through a user study~\cite{zhangPaGELinkPathbasedGraph2023}. However, the authors restricted their comparison to graph designs among themselves only, preventing a comprehensive assessment of the validity of graph explanations against alternative designs.

In this work, we go beyond the conventional algorithmic-oriented evaluation typically employed in the CS field and introduce both qualitative and quantitative user studies to evaluate the validity of graph-based explanations compared to other designs. 


\subsection{Explainable AI in Human Computer Interaction}

It is acknowledged that visualization can provide cognitive support through various mechanisms, \emph{e.g.} helping in pattern discovery, summarizing large volumes of data, or reducing search time~\cite{toryHumanFactorsVisualization2004}. The HCI field extensively studies the influence of design choices, with several works emphasizing their significance in impacting users' understanding and ability to contextualize problems. In particular, numerous studies have explored the impact of graph visualizations and showed the influence of the overall setting on their performance. For instance, several works demonstrated that node-link representations were indeed advantageous over matrix representations, when faced with small graphs and when path-oriented tasks were involved~\cite{ghoniem2004comparison,ghoniem2005readability,okoe2015graphunit,keller2006matrices,okoe2018node}. For other tasks, such as weighted graph comparison~\cite{alperWeightedGraphComparison2013} or suspicious node detection~\cite{mcbride2013efficacy}, matrix representations have proven to be a superior choice over node-link diagrams. More recently, authors in~\cite{digiacomoUserStudyHybrid2021} analyzed hybrid visualizations that combine node-link and matrix representations. Their goal was to provide a comprehensive tool for analyzing real-world networks that are globally sparse but locally dense. They showed that in such configurations, their mixed model overcomes the limitations of using the node-link diagram alone. While all these approaches systematically involve user study evaluations, they are not specific to the explanatory context of recommender systems.

Few works from the HCI field address both graphs and their evaluation as explanations. In an early study, authors in~\cite{o2008peerchooser} carried out a user study to assess the impact of various interactive graph-based representations of a recommender system. They showed that 78\% of users \emph{``felt that the system provided a good explanation of collaborative filtering''}. However, the graph representation itself was not challenged and the study only focused on the layout of the system (profile-based or not) and its interactivity. More recently, authors in~\cite{koukiUserPreferencesHybrid2017} involved graphs to highlight user preferences for \textit{item-based} explanations. However, their graphs consist in concentric circle diagrams or pathways between columns, which we argue do not fully leverage the capabilities of graphs. We draw a similar conclusion for~\cite{ghazimatinPRINCEProvidersideInterpretability2020}, where authors show participants' preference for action-oriented over connection-oriented explanations, but where graphs are limited to the paths they can exhibit. 

In this work, similar to the aforementioned studies, we maintain the evaluation process that involves user studies to assess design performance. However, we deviate from them by integrating our analysis directly into the explanatory context.

\section{Study 1: Qualitative exploration of stakeholders' expectations regarding graph-based explanations}\label{sec:study_1}

We seek to answer \textbf{RQ1} by exploring stakeholders' expectations regarding graph-based explanations for AI-based recommender systems. To achieve this, we conduct a qualitative study involving participants with varying expertise levels in AI systems, employing a think-aloud case study to collect insights and remarks. 

\subsection{Study design}

We interviewed 12 participants. All participants were volunteers, recruited through an email campaign within the university with which the authors are affiliated. We conducted a 30-minute interview with each participant. To ensure data privacy, participants signed a consent form designed and approved in collaboration with the Data Protection Officer. Each interview was then divided into three parts.

In the first part of the session, we assessed the perceived expertise level of participants regarding AI-based systems in general and recommender systems specifically, using a preliminary questionnaire. From these answers, we obtained three distinct groups. The first group consists of \emph{Experts} (4 participants); people who consider themselves \emph{really familiar} with AI-based systems and either have a \emph{precise idea} of what recommender systems are, or developed such algorithms. On the other side of the spectrum, another group includes \emph{non-Experts} (2 participants); these users consider themselves \emph{slightly familiar} with AI-based systems and are only interacting with recommender systems as users, or have a \emph{general idea} of how these systems work. The last group lies in-between these two groups and includes \emph{Insiders} (6 participants); people who are \emph{familiar} with AI-based systems and have a \emph{general idea} of how recommender systems work.

During the second part of the interview, we evaluated participants' perceived understanding of AI-based systems. They answered questions about their anticipated risks for recommender systems, their comprehension of the recommendations when they receive some, and their understanding of the criteria influencing these recommendations.

\begin{figure}
    \centering
    \includegraphics[width=1\linewidth]{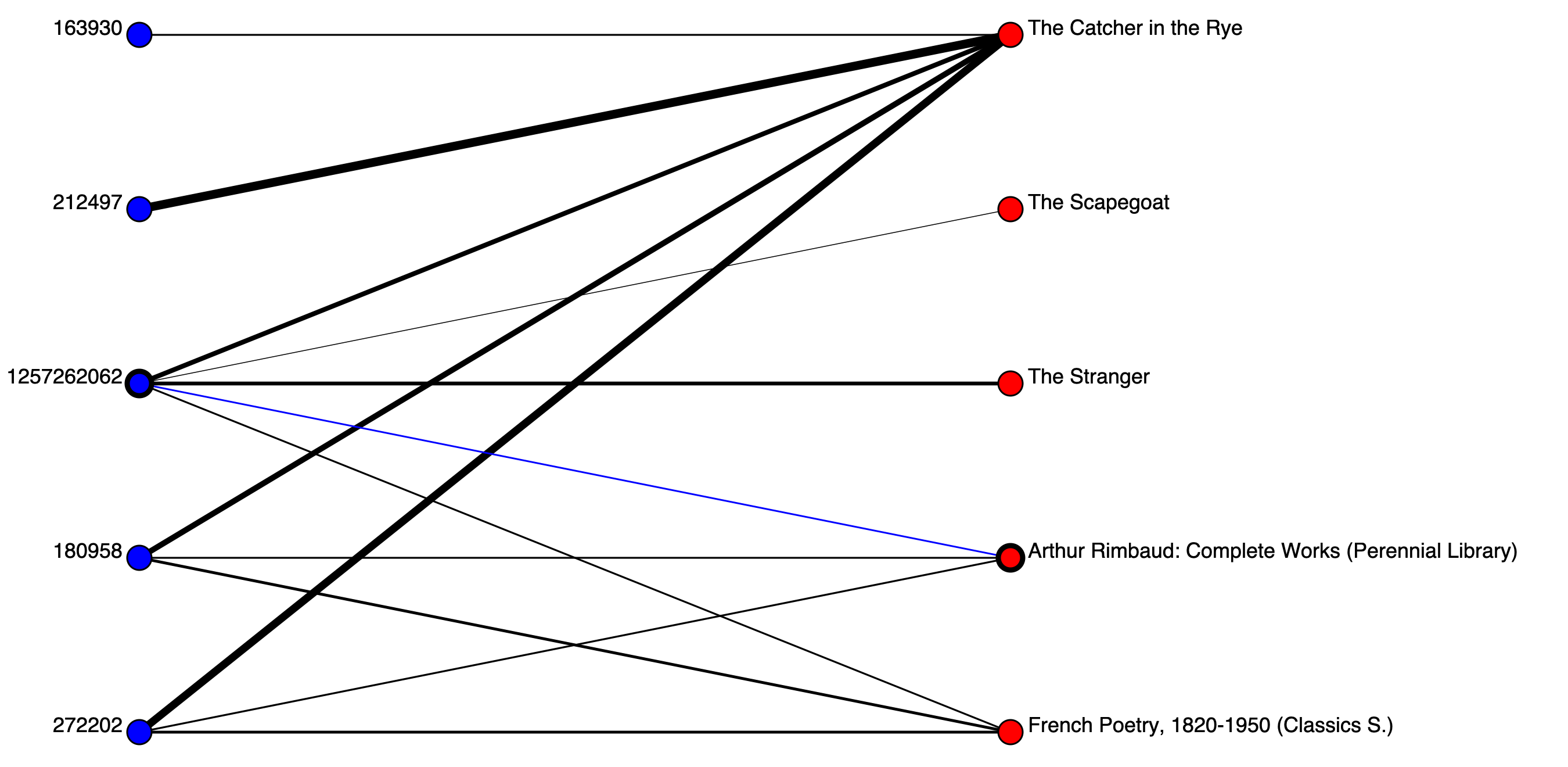}
    \caption{Bipartite graph-based explanation containing two sets of distinct nodes. Blue nodes represent users (along with their identifier) and red nodes represent books (along with their title). A user is linked to a book if the former has read and rated the latter. Thin links denote low ratings, while thick links denote high ratings. The \textcolor{blue}{blue link} corresponds to the system recommendation.}
    \label{fig:graph_based_explanation_50951843}
\end{figure}

Lastly, we introduced a task-oriented think-aloud scenario. In this scenario, we presented participants with an AI-based book recommendation. To increase participant engagement, we collected their book preferences prior to the interview and used this information to create a personalized context incorporated into the AI system, thus making the experiment more realistic. The system recommendation was presented within a graph-based explanation, featuring a subset of the participants' previously read books alongside other users and their readings. This explanation was displayed as a \emph{bipartite graph}, where users were connected to the books they had read, with a distinct link indicating the system recommendation. Lastly, we provided information about book preferences using link weighting (see Figure~\ref{fig:graph_based_explanation_50951843} for an illustration). We did not disclose to the participants the details of the recommendation algorithm or how we had chosen the other visible users. We asked the participants to discuss the relevance of the recommendation to them, their understanding of the elements that led to this recommendation, and what kind of information was missing to better understand the recommendation.

\subsection{Results}

We extracted key insights from the interviews, focusing on participants' understanding of the recommendation through the lens of graph-based explanation. Additionally, we performed an inductive content analysis~\cite{elo2008qualitative} of the notes taken during the interviews, identifying main themes related to participants' desiderata toward graph-based explanation. These results are summarized in Table~\ref{tab:study_1_themes}.


\subsubsection{A high level of perceived understanding, based on three main criteria}
To the question \emph{``To what extent do you understand the suggestions made to you by AI-based recommendation systems?''}, the participants, whatever their level of expertise, said they understood the recommendations in most of the cases. In the few cases when participants find the recommendations unclear, they suggest that their profile diverges from a hypothetical \emph{``basic''} user, or they propose that the algorithm's designers intentionally introduced a mechanism to propose a \emph{``percentage of new things''} unknown to the users.  
When questioned about their understanding of the factors influencing a system to recommend a specific item, participants identified three main criteria. Both \emph{Insider} and \emph{non-Expert} participants reported that the recommendations are primarily built based on user characteristics, such as \emph{``the person's gender, age''} or \emph{``the person's interests''}. Some \emph{Experts} and a few \emph{Insiders} believed that the \emph{``similarity''} between their profile and the profiles of other users takes precedence in the recommendation algorithm's decision. According to this group, the cross-knowledge of a large number of users and their preferences enables the system to suggest new items. Lastly, some participants considered that a recommendation is primarily influenced by their personal history of interactions with the platform. These participants mentioned factors such as \emph{``clicking on an ad''}, the \emph{``frequency of viewing''}, or the \emph{``purchase history''} as elements at the core of the system's suggestions.


\subsubsection{Strong link between context knowledge and recommendation understanding}
When asked \emph{``Do you think this recommendation is relevant? and why?''}, participants unfamiliar with the item suggested by the system either expressed a lack of understanding of the recommendation (\emph{``I do not know this book therefore I don't know if the recommendation is valuable.''}) or asked for additional information about the book's characteristics (\emph{e.g.} author or literary genre) before answering. Conversely, a participant who is familiar with the recommendation or, after reviewing its related characteristics identifies familiar elements (other books by the same author), will immediately deem the recommendation relevant and give it credibility: \emph{``The recommendation seems relevant [...] because I know the author, and it makes me want to read the book.''}. In one case, the participant was familiar with the recommendation, had already read the suggested book, yet found it less relevant. This judgment was based on the perception that the recommendation did not align with the literary genre they prefer. In summary, regardless of their level of expertise, participants tend to draw parallels between the predicted book and their reading history, considering factors such as literary genre, period, or author. The analysis of similarities with other users is only taken into consideration at a later stage and appears to be optional for the positive or negative judgment of the recommendation. In one instance, the recommendation aligns with the participant's literary preferences (known and appreciated author) and is described as relevant, even if the participant does not see the connection with the elements provided in the explanation.

\subsubsection{Graphs as objects modeling both similarities and popularity} Participants' utilization of the graph-based explanation can be characterized along two dimensions. Firstly, \emph{Insiders} and \emph{Experts} follow the links between users and books to highlight similarities among users. These similarities are described through the sharing of literary tastes and represent the manifestation of their a priori understanding of how a recommender system operates. In such cases, the weight (thickness) associated with the links, especially when positive, plays an important role: \emph{``It visually strikes me a bit''}. Secondly, the graph-based explanation leads to an understanding of predictions in terms of popularity, \emph{i.e.} participants focus on the quantity of users who have liked the recommendation. In these cases, the recommendation is seen as a consequence of the enthusiasm surrounding it, rather than its relevance to the individual user.


A few remarks about the use of graphs by participants should be noted. The conclusions drawn regarding user similarities through graph analysis require time (a few seconds), even for \emph{Experts} who, at the end of the interview, considered the graph easy to use. To derive insights about the mechanisms of the system, \emph{non-Experts} tend not to analyze the graph through its links and users, but rather to focus on already-known item characteristics (genre or author). Finally, none of the participants mention graph-specific notions, such as \emph{clique}, \emph{i.e.} fully connected subset elements in a graph or \emph{density}. Intuitively, such elements could have been used to deeply understand subgroups of users that led to the recommendation. 

\subsubsection{Graph-based explanations are considered interpretable but should contain item information}
Participants, regardless of their level of expertise, all agree that it would be useful to have more information about the characteristics of the suggested item, such as literary genre, period, etc. This would enable the creation of \emph{``connections between different books''} and provide a \emph{``general context''} to the decision: \emph{``What I miss is the tool's knowledge about the world.''}, \emph{``I lack an idea of which types of books are similar to each other.''}. To address this, a participant suggests using colors to highlight the proximity between groups of books. Furthermore, we noticed that the needs for additional item information align perfectly with the understanding of the recommendation; more than any other feature, participants require item characteristics that they can rely on to judge the relevance of a given recommendation.

An \emph{Expert} and an \emph{Insider} discussed the value of links representing a weak attraction to items, mentioning that they helped establish a \emph{``contrast''} with other users, thereby enhancing their understanding of \emph{``similarities in their profile with other users.''}. Some \emph{Insiders} evoked adding numerical information on links, rather than playing on thickness. However, the relevance of information about disliked items, \emph{i.e.} thin links in the graph, is not shared by all users: \emph{``I don't really get the thin links in this graph.''}, \emph{``The fact that we have a thick line and a very thin one [...], side by side, confuses me a bit.''}.

\emph{Experts} find graph explanations \emph{``very easily interpretable''} or \emph{``sufficient''} and do not see the need for them to be \emph{``transformed into natural language.''}. One of them interprets them as the visual counterpart to the classic formula found on platforms using recommendation: \emph{``Other users than you, who liked similar things to you, also liked...''}. Moreover, they \emph{``do not expect to have an exhaustive representation''} of the context of the recommendation. Yet, even \emph{Experts} can have trouble distinguishing clear relations with other users; \emph{``I don't see clearly whether I am close or not to other users. [...] I don't fully see the connection between me and other users.''}.

\begin{table*}
\caption{Main themes on graph-based explanation design's desiderata emerging from interviews.}
\label{tab:study_1_themes}
\resizebox{\textwidth}{!}{%
\begin{tabularx}{\textwidth}{XXXX}
\toprule
\thead{\normalsize \textbf{Explanation}} & \thead{\normalsize \textbf{Experts}} & \thead{\normalsize \textbf{Insiders}} & \thead{\normalsize \textbf{non-Experts}} \\
\midrule
\textbf{Global design choice} & Natural language explanation, no need to transform it into natural language, not structured, contrastive (with link thickness), not exhaustive, {[}graphs{]} are really not bad, very easily interpretable, interactions between groups, I don't fully see the connection between me and other users & Links between books, themes of books, adding numbered information to links, The fact that we have a thick line and a very thin one [...], side by side, confuses me a bit. & I don't see what {[}other users{]} liked about {[}an item{]} \\
\midrule
\textbf{Item-based vs. user-based} & You like this book and there is this other similar book, connections between different books vs. I don't see if I'm close to other users, I miss visualization of my similarity with other users & How to find the book characteristics?, I lack an idea of which types of books are similar to each other, thicker links if books are of the same kind, links between books and themes vs. both this user and I did not like this book & {[}About books{]} Do you consider classics? Where do authors come from? Did the books receive prizes? \\
\bottomrule
\end{tabularx}%
}
\end{table*}

\section{Study 2: Influence of explanation design}\label{sec:study_2}

To address \textbf{RQ2}, \textbf{RQ3} and \textbf{RQ4}, we conduct a qualitative user study in which we investigate the influence of explanation design, among them a graph-based explanation, on various AI-based recommender system users. An example of our interface is shown in Figure~\ref{fig:questionnaire} in the Appendix.

\subsection{Study design}

\subsubsection{Recommender system}

For this purpose, we use the setup introduced in our qualitative experiment (see Section~\ref{sec:study_1}): we build an AI-based book recommender system which goal is to provide the participant with a book suggestion given pre-selected reading preferences. We did not share the functioning details of the recommender system with the participants, but specified that the recommendation they were given was AI-based.

\subsubsection{Explanation design}

\begin{figure*}
  \begin{subfigure}[t]{0.46\textwidth}
    \centering
    \includegraphics[width=\linewidth]{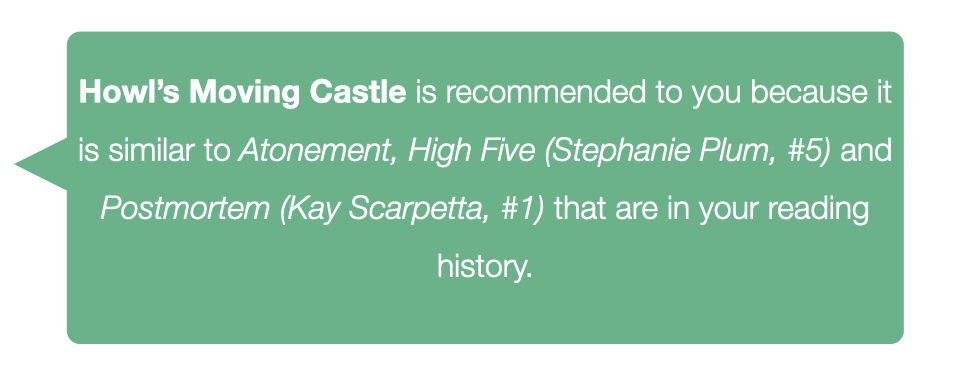}
    \caption{Text-based explanation. Item-based wording is chosen accordingly with results from~\cite{kouki2019personalized}.}
  \end{subfigure}%
  \hfill
  \begin{subfigure}[t]{0.53\textwidth}
    \centering
    \includegraphics[width=\linewidth]{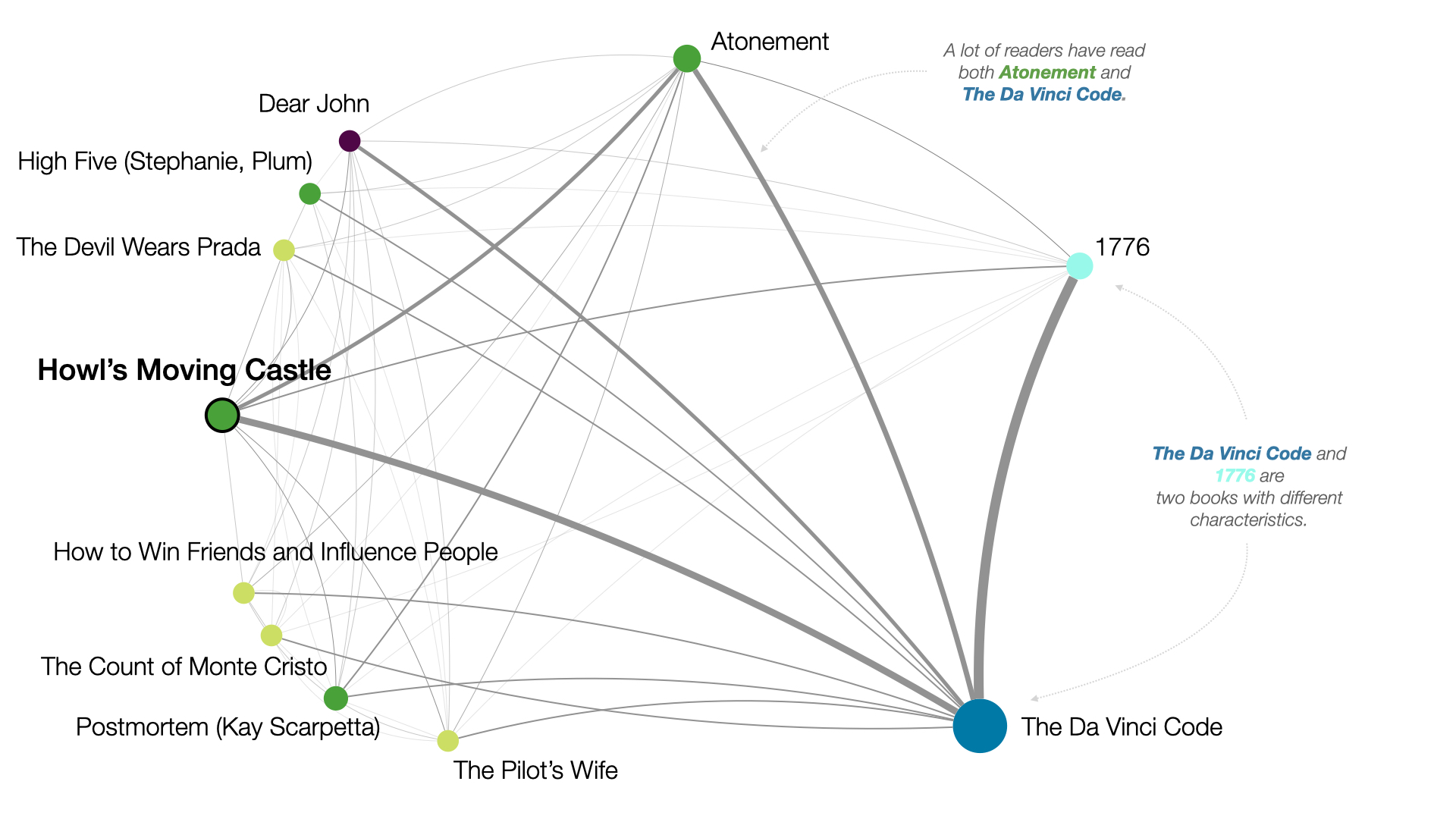}
    \caption{Graph-based explanation. Nodes represent the books read by the user, except for the black-circled node which corresponds to the recommendation. Node colors denote similarities between book characteristics. A link between two books exists if they both have been read by some user. Thickness of links stands for the number of common readers for two books.}
  \end{subfigure}

  \begin{subfigure}{\textwidth}
    \centering
    \includegraphics[width=0.9\linewidth]{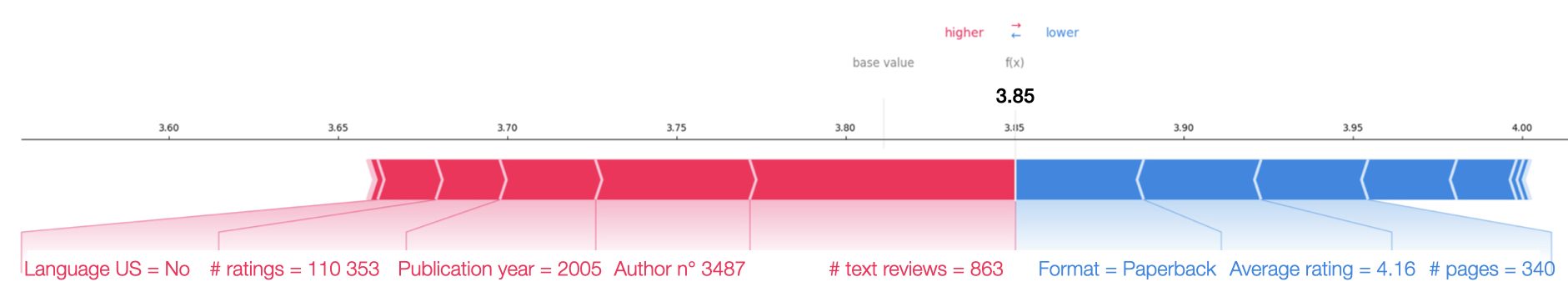}
    \caption{SHAP-based explanation~\cite{lundberg2017unified}. Recommendation is explained through feature importance.}
  \end{subfigure}
  
  \caption{Explanation designs used in the second user study.}
  \label{fig:explanation_design_example}
\end{figure*}

We introduced three different explanation designs, that were detailed to each participant through a small paragraph, as well as a reading key. We used the well-known \emph{SHAP}~\cite{lundberg2017unified} approach to design a feature importance oriented explanation design. This choice stems from the popularity of this method in the Explainable AI (XAI) community as well as for its tight bounds with the users' expectations obtained from our first study, \emph{i.e.} book characteristics are considered important for the quality of the explanation. We also presented a \emph{Text} explanation, containing concise item-based sentences stating \emph{why} this recommendation was made. This choice was influenced by recent studies indicating that text explanations, specifically with item-based wording, were perceived more persuasive than other visual formats~\cite{kouki2019personalized} and were effective in providing the user with personalized feeling, which is correlated with trust in the system~\cite{zhang2018exploring}. Finally, we improved the \emph{Graph} design introduced in Figure~\ref{fig:graph_based_explanation_50951843} based on participants' feedback from our qualitative study. Specifically, we focused on item characteristics by projecting the initial bipartite graph into an item-oriented one. In this representation, each book is a node, and edges connect books that share common readers in our database. Additionally, we incorporated information about book characteristics (literary genre, author, publication year, etc) through node colors. Consequently, this final graph representation aligns more closely with users' expectations. We illustrate these designs in Figure~\ref{fig:explanation_design_example}.

\subsubsection{Experimental conditions}

Overall, 66 participants were recruited through two channels: (i) students, doctoral students and researchers from the author's university and affiliated laboratories, and (ii) users of the Reddit forum \emph{``SampleSize''}\footnote{\url{https://www.reddit.com/r/SampleSize/}} dedicated to user studies. All participants were asked to fill out an online form divided into two parts. First, as in the previous study, participants answered preliminary questions about their perceived expertise level toward AI-based systems in general and AI-based recommender systems in particular. We ended up with 11 \emph{non-Experts} (16.7\%), 32 \emph{Insiders} (48.5\%) and 23 \emph{Experts} (34.8\%). We then asked each participant to choose their favorite book selection from among 5 sets of 10 books and used our AI-based recommender system to suggest a new book. Each participant was shown all explanation designs for this recommendation (in a random order), and for each design, they were asked to answer a set of questions related to our evaluation measures. Finally, we collected participants' design preferences by explicitly asking them to rank the three designs. The whole study lasted around 10 minutes.

In summary, this study is constructed using a Mixed Factorial design, involving the participants' \emph{expertise level} as a between-subject variable and the \emph{explanation design} as a within subject variable. Therefore, in the following analysis, we assess three effects: (i) the influence of the within-subject variable on the measures, (ii) the influence of the between-subject variable on the measures, and (iii) the effect of the interaction between the two factors on the measures.

\subsubsection{Evaluation constructs}

The same set of questions was used for each explanation design. These questions are built according to previous literature on explanation evaluation within AI-based frameworks~\cite{o2015empirical,hoffman2018metrics,purificatoEvaluatingExplainableInterfaces,bertrand2023questioning} and are shaped to fit our specific experimental design. All questions have the same structure; given an explanation design, we ask the participant \emph{``Please evaluate your level of agreement with the following sentences.''}. Each participant answers using a 5-point Likert scale ranging from \emph{Strongly disagree} to \emph{Strongly agree}. We organized the study around four global constructs (detailed below). We verified the internal validity of the questionnaire by computing Pearson correlations between answers within each construct and the total of each construct (at this stage we discarded results from one question), and its internal reliability using McDonald's $\omega$~\cite{mcdonald2013test}. Detailed questions and corresponding categories are provided in Table~\ref{tab:study_2_questions}.

\textbf{Subjective understanding.} We measured subjective understanding by asking participants if they understood the recommendation made by the system or if they were able to derive insights about its internal mechanics.

\textbf{Objective understanding.} We measured objective understanding of the recommendations by asking participants about the features used by the system to provide the recommendation. These questions were specifically designed for the study but encompass different recommender systems elements that have been emphasized in our first study as well as in previous studies from the literature~\cite{kouki2019personalized,bertrand2023questioning}, \emph{e.g.} item-based influence, user-based influence or self-historical information influence.

\textbf{Usability.} Usability was measured by asking participants how difficult the explanation was to read and understand. We also asked participants about how easily other people would learn to read this explanation~\cite{alabi2022measuring}.

\textbf{Curiosity.} We measured the curiosity induced by the explanation design using questions adapted from~\cite{o2015empirical,hoffman2018metrics}. We asked the participants if they were curious to know why the recommender system did not provide another suggestion, and if the recommendation incited curiosity.

\begin{table*}
\caption{Questions asked to participants for each explanation design, with the $p$-value corresponding to the validity test (Person's correlation between each question and its corresponding global construct) and the McDonald's $\omega$ score.}
\label{tab:study_2_questions}
\resizebox{\textwidth}{!}{%
\begin{tabularx}{\textwidth}{lccX}
\toprule
\textbf{Construct} & \textbf{Reliability $\omega$} & \textbf{Validity} & \multicolumn{1}{c}{\textbf{Question}} \\ 
\toprule
\multirow{2}{*}{Subjective understanding} & \multirow{2}{*}{84.4} & \multirow{1}{*}{$p<.001$}  & I understand why this recommendation was made to me. \\
 & & \multirow{1}{*}{$p<.001$} & I can figure out the internal mechanics of the recommender system. \\ 
 \midrule
\multirow{3}{*}{Objective understanding} & \multirow{3}{*}{63.5} & \multirow{1}{*}{$p<.001$}  & I believe this book was recommended to me because it is similar (literary genre, author, etc.) to the books I chose previously. \\
 & & \multirow{1}{*}{$p<.001$}  & The fact that I share preferences with other readers only counted a little by the algorithm. \\
 & & \multirow{1}{*}{$p<.001$} & I believe that this book was recommended to me due to its popularity among other readers. \\ 
 \midrule
 \multirow{2}{*}{Usability} & \multirow{2}{*}{48.2} &  \multirow{1}{*}{$p<.001$} & I would imagine that most people would learn to read this explanation very quickly. \\
 &  & \multirow{1}{*}{$p<.001$} & I think the recommender system is interpretable. \\ 
 \midrule
 \multirow{1}{*}{\emph{Usability (not used)}} & \multirow{1}{*}{-} & \multirow{1}{*}{$p>.05$} & \emph{I found it difficult to read and understand the recommendation context.}\\
 \midrule
\multirow{2}{*}{Curiosity (0.510)} & \multirow{2}{*}{60.5} & \multirow{1}{*}{$p<.001$} & I am curious about why the recommender system did not make other decisions. \\
 & & \multirow{1}{*}{$p<.001$} & The recommended book incited my curiosity. \\ 
 \bottomrule
\end{tabularx}%
}
\end{table*}


\subsection{Results}
\begin{figure*}
    \centering
    \begin{subfigure}[b]{0.33\textwidth}
        \includegraphics[width=\linewidth]{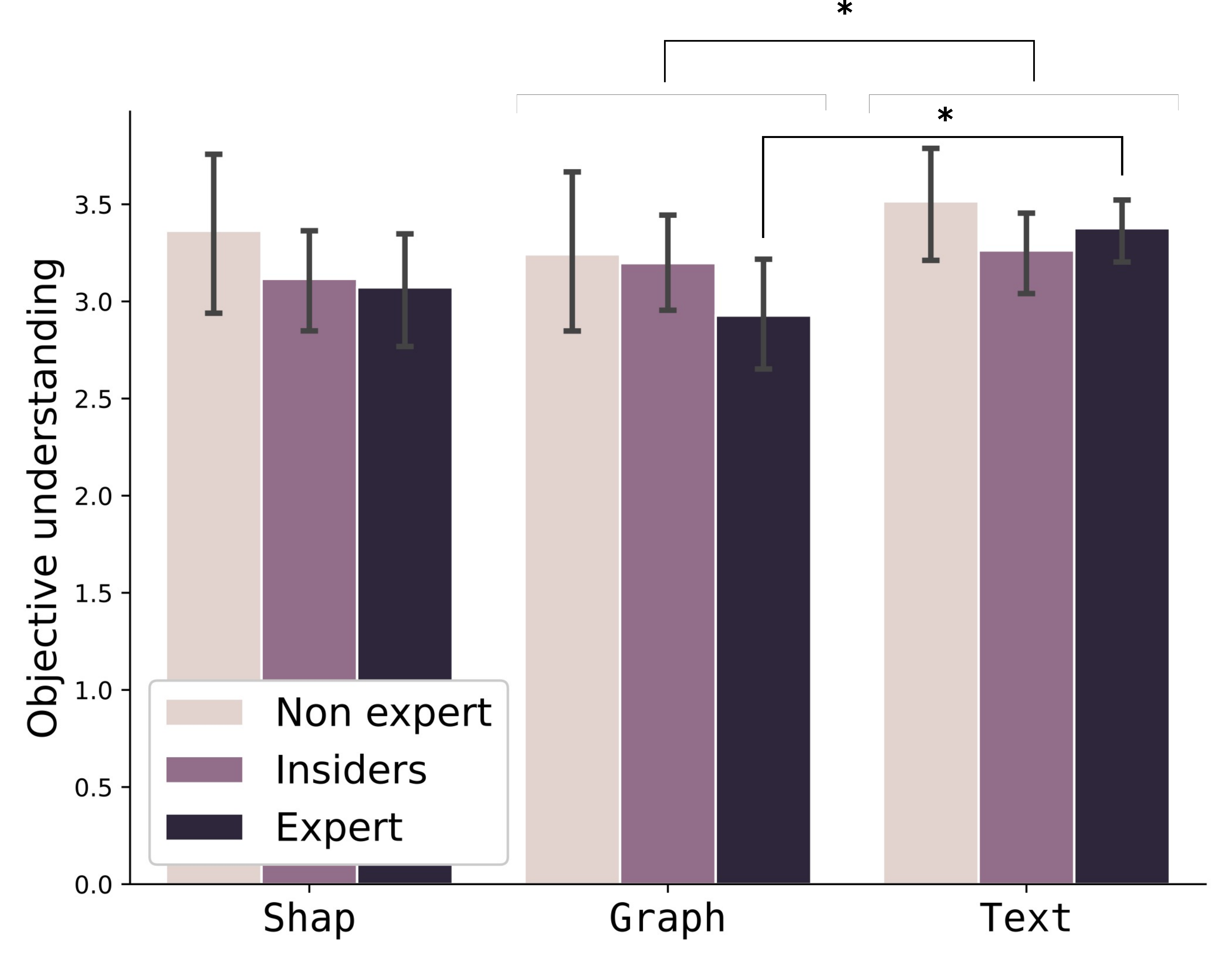}
        \caption{Objective understanding}
        \label{fig:sub1}
    \end{subfigure}
    \hspace{0.02\textwidth} 
    \begin{subfigure}[b]{0.33\textwidth}
        \includegraphics[width=\linewidth]{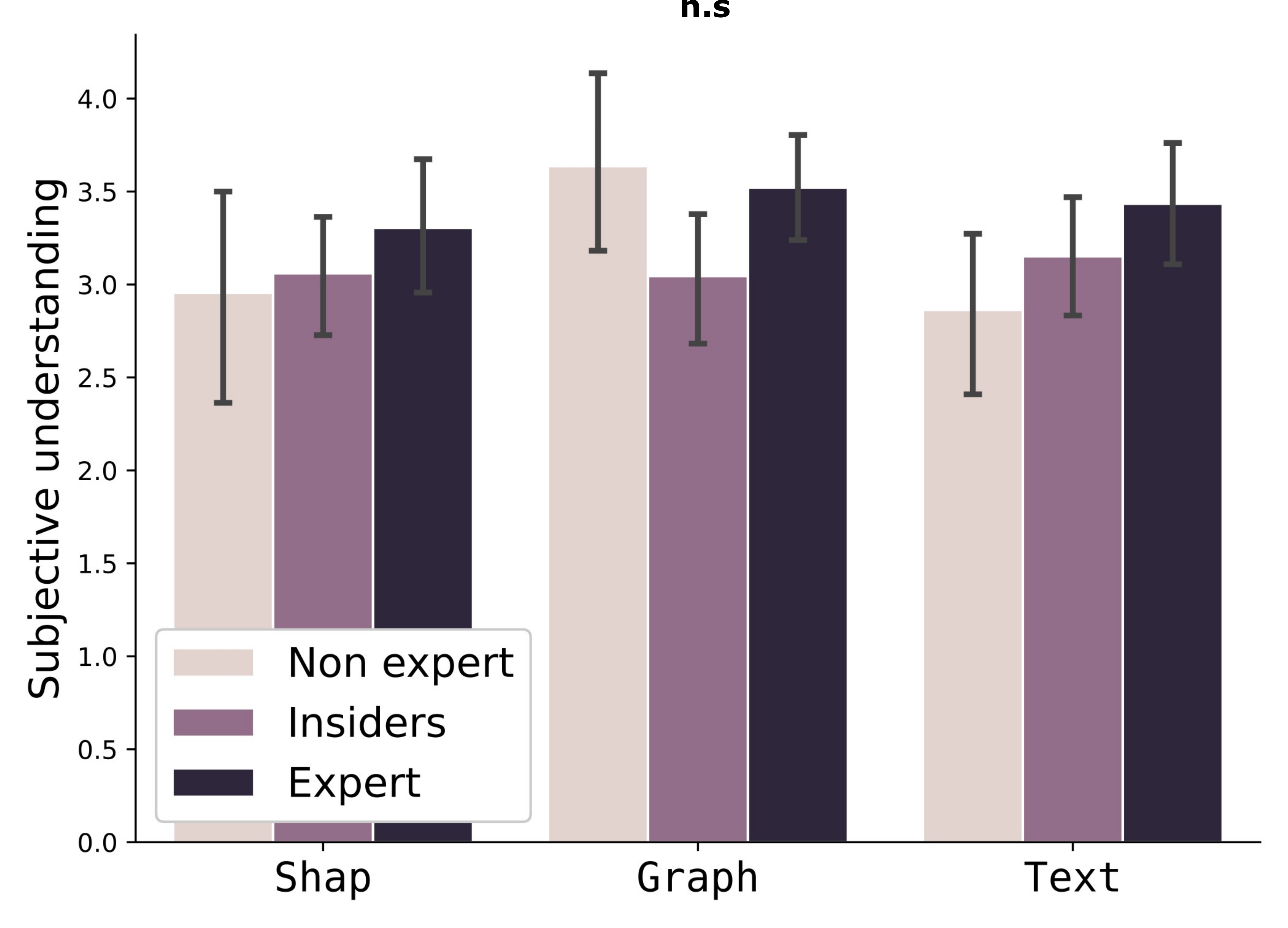}
        \caption{Subjective understanding}
        \label{fig:sub2}
    \end{subfigure}
    \vspace{1em} 
    \begin{subfigure}[b]{0.33\textwidth}
        \includegraphics[width=\linewidth]{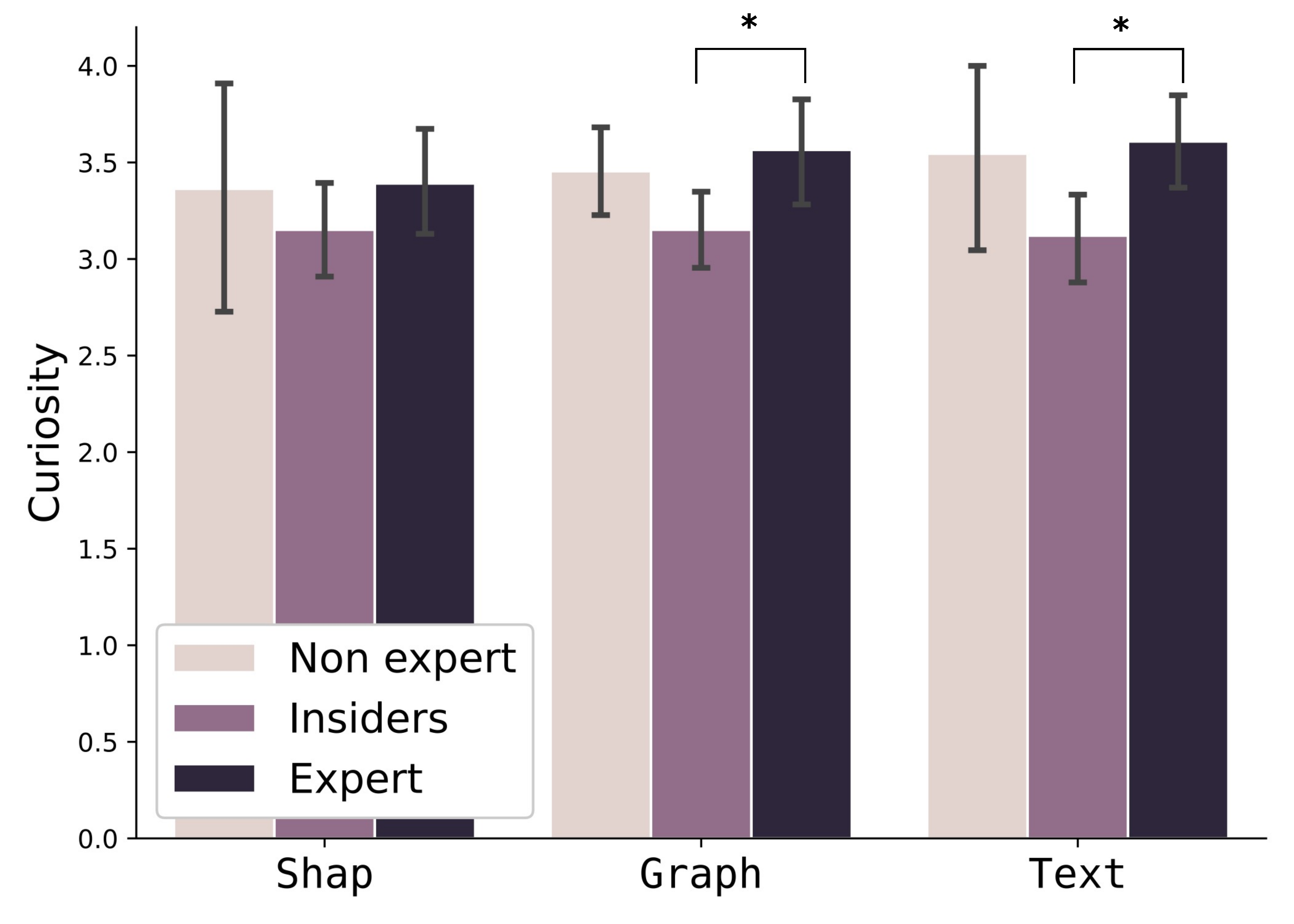}
        \caption{Curiosity}
        \label{fig:sub3}
    \end{subfigure}
    \hspace{0.02\textwidth} 
    \begin{subfigure}[b]{0.33\textwidth}
        \includegraphics[width=\linewidth]{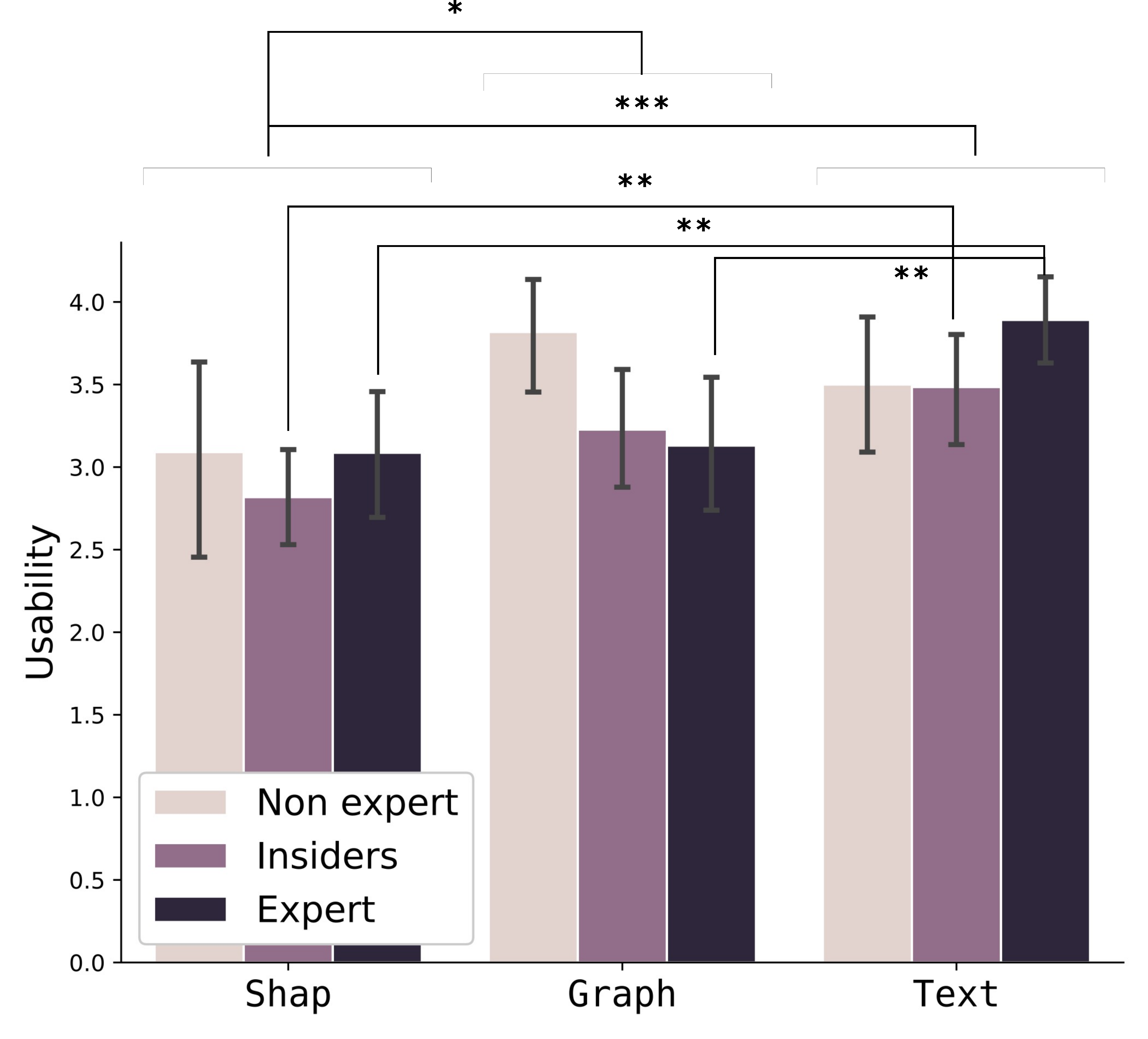}
        \caption{Usability}
        \label{fig:sub4}
    \end{subfigure}
    \caption{Results for quantitative study. Vertical bars are confidence intervals at $95\%$. Significance levels are reported as follows: $***=p \leq .001, **=p \leq .01, *=p \leq .05$,  $\cdot=p \leq .07$ and n.s. non significant. \emph{Reading key: Objective understanding (Figure \ref{fig:sub1}) was statistically significantly higher ($p \leq .05$) when participants used textual explanation rather than graph explanation.}}
    \label{fig:study_2_results}
\end{figure*}

We used a repeated-measure analysis of variance (RM-ANOVA) to analyze the collected measures for each participant. Each measure consisted in the average of the ratings for the corresponding questions. All measures passed the sphericity assumption (constant variance across repeated measure), either using the traditional Mauchly's test or after applying Greenhouse-Geisser correction. We confirmed the homogeneity of variance for all levels of the repeated measures using Levene's test. When significant effects were observed, we conducted a post-hoc Bonferroni test for pairwise comparisons. Results are summarized in Figure~\ref{fig:study_2_results}.

\subsubsection{Text rather than graph design for higher objective understanding}

We found a statistically significant main effect of explanation design on participants' objective understanding ($p=.043$). Post-hoc analysis with a Bonferroni adjustment revealed that text-based design led to significantly higher objective understanding than graph-based design ($p=.046$) (see Figure~\ref{fig:sub1}), but that there was no statistically significant difference for the levels of objective understanding between graph and SHAP-based designs, nor between text and SHAP-based designs. The main effect of the expertise level on objective understanding was not statistically significant ($p=.424$), \emph{i.e.}, if we ignore the explanation design being evaluated, \emph{Experts, Insiders} and \emph{non-Experts} gave similar ratings considering the objective understanding measure. However, within the \emph{Expert} group, objective understanding was statistically significantly higher for text-based design than for graph-based design ($p=.022$).

\subsubsection{Graph and text designs increase usability. }

There was a statistically significant main effect of the explanation design on usability ($p<.001$), but no statistically significant influence of the level of expertise on this measure ($p=.398$). Post-hoc tests with a Bonferroni adjustment revealed that SHAP-based design led to statistically significantly lower usability compared to graph ($p=.043$) or text-based ($p<.001$) designs (see Figure~\ref{fig:sub4}). However, we did not find any statistically significant difference between graph and text-based designs ($p=.418$). Within expertise levels, SHAP-based design led to statistically significantly lower level of usability than text-based design, for both \emph{Insiders} ($p=.007$) and \emph{Experts} ($p=.006$). Among \emph{Experts}, usability was also statistically significantly higher for text-based design than for graph-based design ($p=.008$).

\subsubsection{Higher expertise increases curiosity for graph and text designs.}

Our analysis determined that there was no statistically significant effect of explanation design on participants' curiosity ($p=.514$). However, the expertise level had a statistically significant influence on curiosity ($p=.016$). Post-hoc tests with a Bonferroni adjustment revealed that being \emph{Insider} led to significantly lower levels of curiosity compared to being \emph{Expert} ($p=.02$). Specifically, when faced with graph-based designs, \emph{Experts} showed statistically significantly higher levels of curiosity than \emph{Insiders} ($p=.041$). Similarly, when faced with text-based designs, \emph{Experts} were significantly more curious than \emph{Insiders} ($p=.032$) (see Figure~\ref{fig:sub3}). However, we did not find any statistically significant difference in curiosity levels between \emph{non-Experts} and other users. Furthermore, we did not find any statistically significant difference in ratings between participants when using SHAP-based design.

\subsubsection{No effect of explanation design or expertise level on subjective understanding}

We did not observe any statistically significant influence of explanation design on participants' subjective understanding ($p=.104$). Similarly, the expertise level of participants did not have any statistically significant influence on their subjective understanding ($p=.185$) (see Figure~\ref{fig:sub2}).

\subsubsection{Graph-based design is preferred, regardless of the expertise level}

\begin{figure}
    \centering
    \includegraphics[width=1\linewidth]{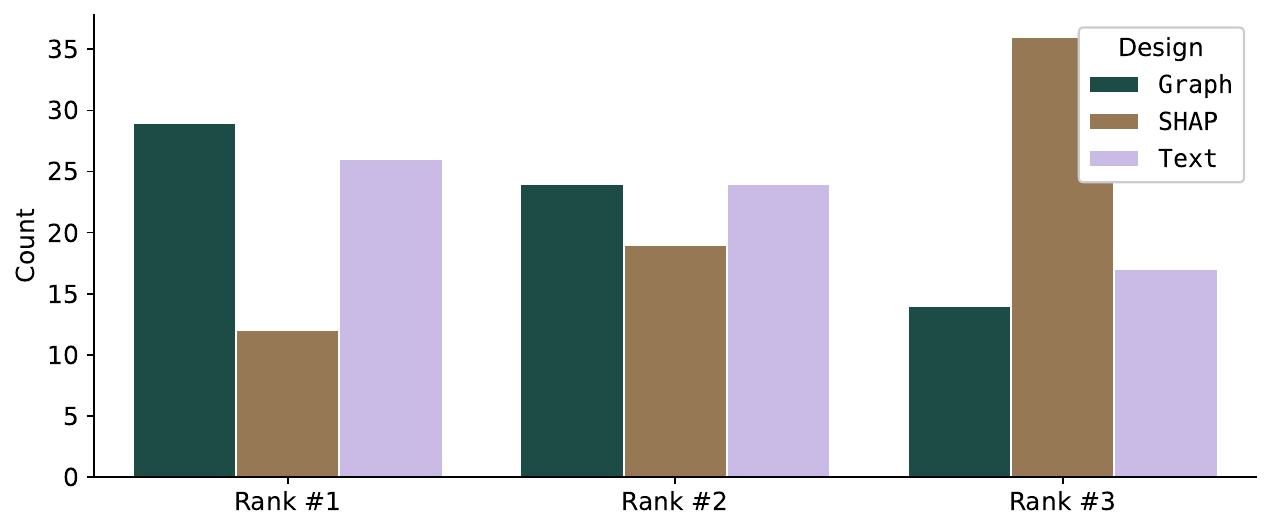}
    \caption{Number of occurrences of each design at each ranking position. \emph{Reading key: Graph-based design has been ranked 29 times in \#1 position, 24 times in \#2 position and 13 times in \#3 position.}}
    \label{fig:design_order}
\end{figure}

At the end of the questionnaire, participants were asked to rank explanation designs according to their preferences. A Chi-Square Goodness of Fit Test was performed to determine whether the design preferences were equally distributed across the 6 possible outcomes. Our results revealed that the obtained proportions significantly differed ($p=.015$). We display in Figure~\ref{fig:design_order} the number of occurrences of each design at each ranking position. We show that participants preferred graph-based design more frequently than other designs. When graph-based design was not ranked in first position, it was ranked in second position more often than in third position. The second best-ranked design was text-based explanation. SHAP-based explanation was ranked in last position more than 35 times. Finally, we explored the relationship between participants' expertise level and design preference. We ran a Chi-Square test and did not find any statistically significant evidence of correlation between the two variables ($p=.841$).

\section{Discussion}

\subsection{Graph vs. textual explanations}

According to Miller~\cite{miller2019explanation}, a ``good'' explanation should be \emph{contrastive, selected, dialogic} and \emph{causal}. We argued that graphs inherently model causal and contrastive relationships between entities through their node-link structure. They achieve the \emph{selected} criterion by potentially representing a subset of the original data, \emph{e.g.} in the vicinity of a recommender system's prediction. Finally, the node or link attributes they may possess contribute to them being considered dialogic. Consequently, we aligned with previous studies that acknowledged the strong expressive power of graphs and the fact that they were especially well-suited for applications like recommender systems~\cite{afchar2022explainability}.

Initial findings from our qualitative study affirmed these statements and emphasized the potential effectiveness of graph-based explanations. With the exception of one \emph{Expert} who preferred textual explanations, participants found graph-based design \emph{``very easily interpretable''} and highlighted the importance of showcasing \emph{``relations between groups''}. This was further confirmed quantitatively; when asked for their preferred explanation type, the majority of participants in our second study favored graph-based explanation over text or SHAP designs (see Figure~\ref{fig:design_order}). Specifically, users expressed a preference for item-centric graph explanations over user-centric ones, aligning with results from Kouki \emph{et al.}'s work~\cite{koukiUserPreferencesHybrid2017,kouki2019personalized}, which compared textual and visual explanations, although not specifically graphs.

However, despite their expressed preferences for graph-based design, our second study revealed that participants had significantly higher levels of objective understanding and usability when using textual designs. Such contrast between expressed desiderata and measured performance aligns with previous research~\cite{szymanski2021visual,buccinca2021trust} where authors emphasized users' preference for visual explanations (though not specifically graphs) over texts, despite their poorer performance when using the former kind. A plausible explanation for this tendency may be grounded in the findings of a recent perception study~\cite{muth2015stream} in which authors showed that interest and attractiveness of a stimuli can be well predicted by the complexity of this stimuli; in our case, the visual complexity of our graph explanation, compared to textual explanation could encourage participants to prefer this design over others. More recent research~\cite{carbon2018power} explored this direction in the context of graph designs and further confirmed the close relationship between complex visual structure and interest. 

Overall, our findings suggest that depending solely on stakeholders' expressed desiderata for crafting graph-based explanation designs may not be sufficient. Considering both their preferences and measured performance when using such explanations could provide a more comprehensive understanding of what constitutes a good explanation.

\subsection{Impact of expertise level}

Despite its enhanced design tailored to the needs of users with various levels of expertise, our graph explanation did not lead to significantly higher ratings on the different measures. Nevertheless, noteworthy insights can be derived from this visual design. For example, we showed that high expertise level positively influenced participants' objective understanding when facing textual explanation vs. graph explanation, but we did not reveal any difference in ratings for \emph{non-Experts}. This contrasts with the findings in~\cite{szymanski2021visual} showing how lay users performed better with textual explanations vs. visual explanations. While authors evoke the plausible effects of \emph{confirmation bias, i.e.} favoring elements confirming preconceived ideas, on lay users to explain their results, we did not notice such behavior in our qualitative study. An interpretation of our results could be that \emph{Experts} may be more inclined than \emph{non-Experts} to have preconceived notions, or \emph{false narratives} about the internal functioning of the system, a behavior that has been particularly observed for experienced users by previous works~\cite{tversky1974judgment,kaur2020interpreting}. These \emph{a priori} may be more contradicted by complex designs, such as SHAP or graph, compared to relatively vague explanations like the textual design we proposed. 

We also demonstrated higher curiosity levels for \emph{Experts} compared to \emph{Insiders}, when using graph and text designs, which aligns with previous research~\cite{niu2021luckyfind} stating that curiosity is strongly associated with intrinsic interest and motivation. We hypothesize that \emph{Expert} participants are more engaged in the proposed tasks due to their interest in the topic. However, this tendency is not observed with SHAP-based explanation, warranting further analysis for comprehensive understanding. 

Our analysis finally revealed that both \emph{Insiders} and \emph{Experts} considered text-based design more usable than SHAP-based design, but only the most experienced participants found text-based explanations more usable than graph-based ones. This could be attributed to users with moderate expertise being more persuadable by graphical design and experiencing some kind of \emph{over-trust}~\cite{eiband2019impact}. Or it may be the result of \emph{Experts} being more inclined to over-analyze graph explanations to align with their knowledge of recommender systems and network layouts. Additionally, we did not find any statistically significant difference between levels of expertise when measuring graph usability. This observation aligns with previous research in the biological field~\cite{singh2018crowdlayout}, showing that novice users could create and evaluate graph layouts as effectively as domain experts.

In summary, we highlighted how graph-based explanations were subject to different cognitive biases according to the epxertise level of the end-users. To address this limit, future work could involve further inclusion of \emph{cognitive} factors~\cite{shin2019homo} in explanations to further enhance perceived designs.

\subsection{Limitations}


In our quantitative study, we compelled participants to choose from pre-defined user profiles based on a selection of representative books. This action further determined the algorithm recommendation and consequently the content of each explanation. While this approach allows us to use pre-computed answers, which significantly simplifies the questionnaire procedure, we acknowledge that it might hinder participants from fully recognizing their tastes within the recommender system's choices, potentially influencing their design preferences. To address this concern, we verified that the user profile chosen by participants was not correlated with their design preferences. For this purpose, we ran a Chi-Square test between the two variables and the results revealed that there is not enough evidence to suggest an association between the selected user profile and design preferences ($p=.549$).

In this work, we did not impose any time constraint on participants during their analysis of design, and the only time indication provided was that the questionnaire should take around 10 minutes to complete. Consequently, participants had sufficient time to explore and analyze each recommendation, which may not entirely replicate a real-world scenario where end users require an explanation to make a decision. Future research aimed at assessing the generalizability of our findings could involve time-constrained tasks tailored for specific contexts.

More generally, the results presented in this work are derived from data related to book recommendations. This simplistic use case facilitated the building of the study since it mitigated potential ethical concerns. While our results may have applicability in other domains, additional experiments would be necessary to confirm this.

\section{Conclusion}

In this work, we have conducted a qualitative study to understand the needs for graph-based explanation designs expressed by users with various levels of expertise. From these results, we built an enhanced graph-based design emphasizing several key recommendations such as a content-based approach obtained through bipartite graph projection or the focus on item similarity rather than user-similarity. We then conducted a quantitative study in which we investigated the influence of SHAP, textual and graph-based explanation designs regarding four constructs: objective understanding, subjective understanding, curiosity and usability. Our results revealed that text-based explanations significantly improved objective understanding to graph-based explanation, and that it was specifically marked for users with a higher level of expertise. On the other hand, we did not find any statistically significant evidence of influence of explanation design on subjective understanding. We also showed that both graph and text explanations were considered more usable than SHAP-based design, and that it was particularly true for users with middle to high level of expertise. Moreover, we highlighted the influence of expertise level on curiosity by showing how expert users were more curious than insiders toward graph and textual explanations. Lastly, we emphasized the discrepancy between participants' expressed preferences for graph-based explanations during qualitative interviews and further confirmed quantitatively, and their higher ratings regarding understanding or usability for textual designs. This outcome suggests that solely fulfilling stakeholders' desiderata may not suffice to achieve ``good'' explanations. Crafting hybrid designs achieving a balance between social expectation and effective downstream performance emerges as a significant challenge.

\section{Ethical concerns}

To ensure participants' privacy, the qualitative study was designed and approved in collaboration with the Data Protection Officer (DPO) of the authors' affiliated school. Anonymized versions of interview notes for data analysis are securely stored, with access restricted to the authors. We did not collected any personal information for the quantitative study.

\begin{acks}
\end{acks}

\bibliographystyle{ACM-Reference-Format}
\bibliography{main.bbl}

\appendix

\begin{figure*}[b]
  \begin{subfigure}[c]{0.5\linewidth}
    \centering
    \includegraphics[width=\linewidth]{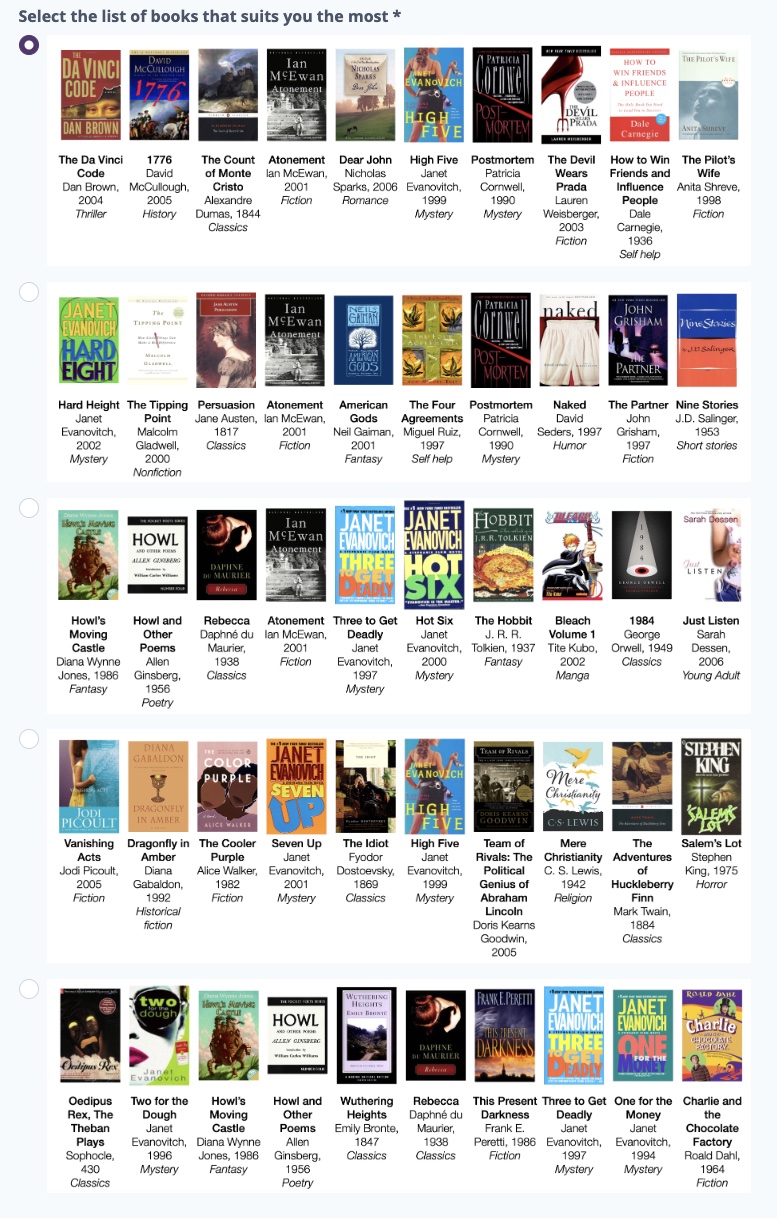}
    \caption{Participants are asked to select a user profile based on a selection of books.}
  \end{subfigure}%
  \begin{subfigure}[]{0.5\linewidth}
    \begin{subfigure}[]{\linewidth}
      \centering
      \includegraphics[width=\linewidth]{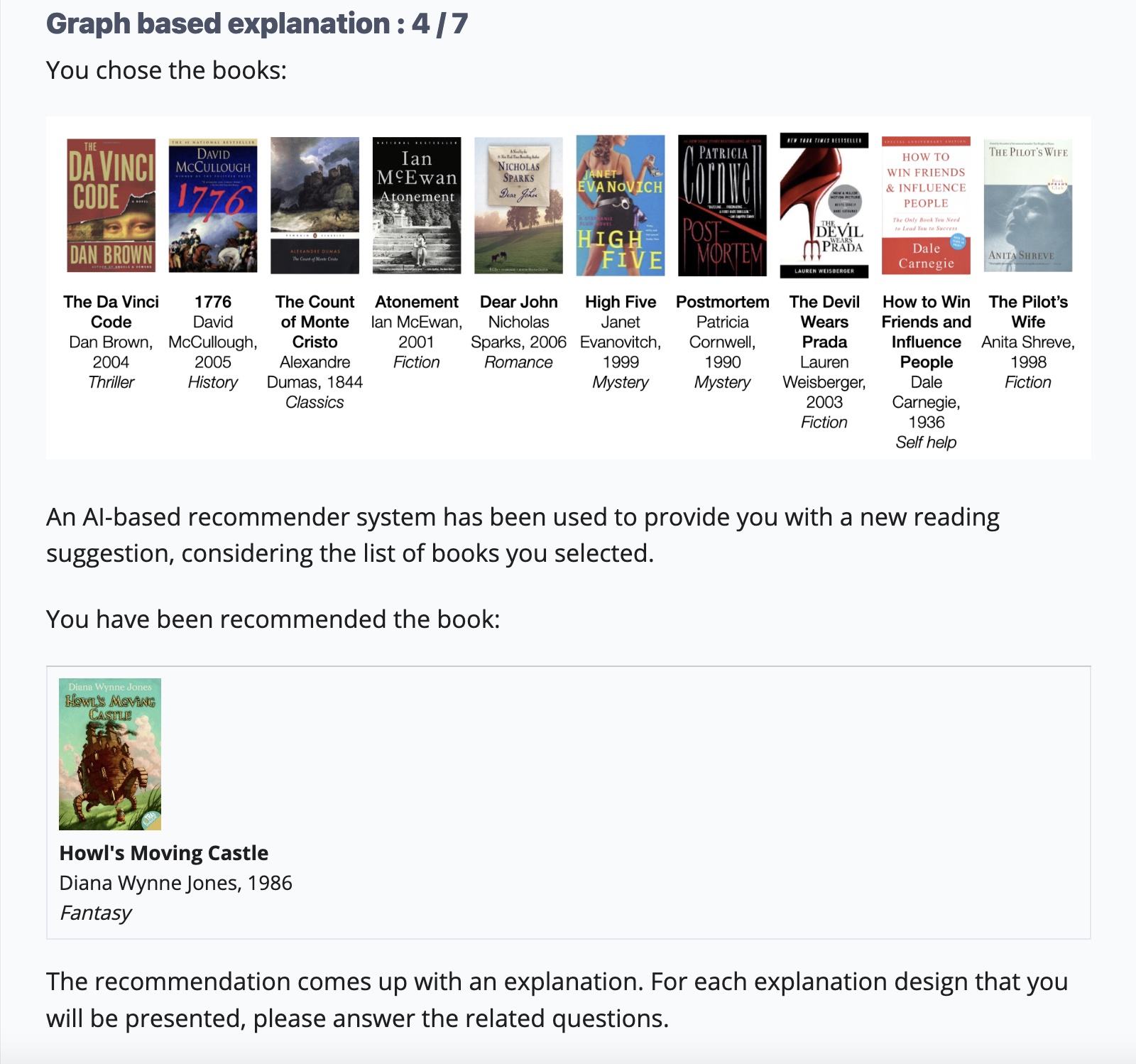}
      \caption{Participants are reminded of their profile choice and are recommended a new book.}
    \end{subfigure}
    \vspace{1em} 
    \begin{subfigure}[]{\linewidth}
      \centering
      \includegraphics[width=\linewidth]{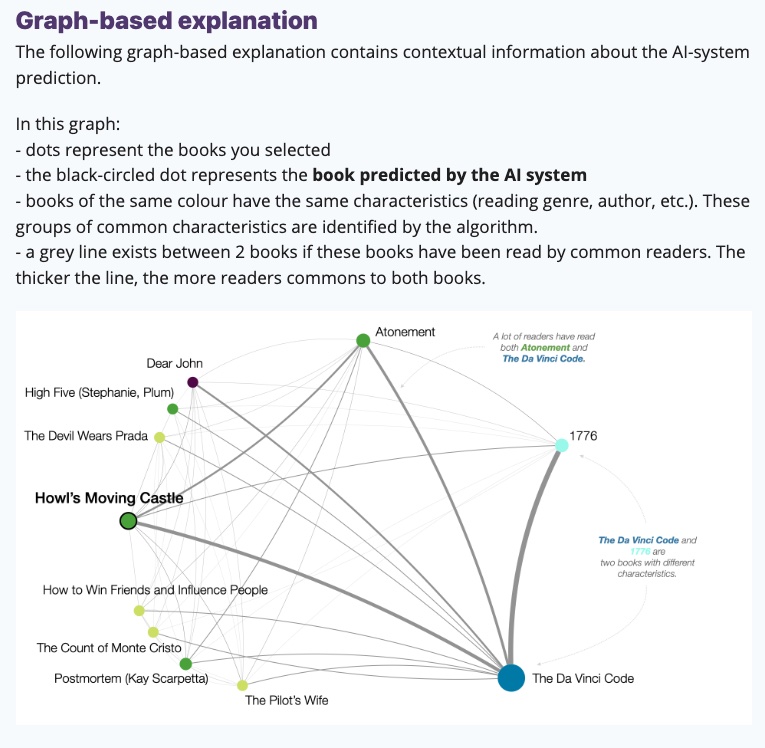}
      \caption{Example of explanation design description (here graph-based design.)}
    \end{subfigure}
  \end{subfigure}
  \caption{Example of a participant's progression through the questionnaire.}
  \label{fig:questionnaire}
\end{figure*}

\end{document}